\pdfoutput=1

\documentclass[10pt, a4paper]{article}
\usepackage{lrec2022} 
\usepackage{multibib}
\newcites{languageresource}{Language Resources}
\usepackage{graphicx}
\usepackage{tabularx}
\usepackage{soul}
\usepackage{titlesec}
\titleformat{\section}{\normalfont\large\bfseries\center}{\thesection.}{1em}{}
\titleformat{\subsection}{\normalfont\SmallTitleFont\bfseries\raggedright}{\thesubsection.}{1em}{}
\titleformat{\subsubsection}{\normalfont\normalsize\bfseries\raggedright}{\thesubsubsection.}{1em}{}
\renewcommand\thesection{\arabic{section}}
\renewcommand\thesubsection{\thesection.\arabic{subsection}}
\renewcommand\thesubsubsection{\thesubsection.\arabic{subsubsection}}

\usepackage{epstopdf}
\usepackage[utf8]{inputenc}

\PassOptionsToPackage{hyphens}{url}
\usepackage{hyperref}
\usepackage{xstring}

\usepackage{color}

\usepackage{enumitem}
\usepackage{booktabs}
\title{Multilingual Open Text Release 1:\\ 
\textbf{Public Domain News in 44 Languages}}

\name{Chester Palen-Michel$^*$, June Kim$^*$, Constantine Lignos}

\address{Michtom School of Computer Science\\
         Brandeis University\\
         \{cpalenmichel, junekim, lignos\}@brandeis.edu\\}

\abstract{
We present Multilingual Open Text (MOT), a new multilingual corpus containing text in 44 languages, many of which have limited existing text resources for natural language processing.
The first release of the corpus contains over 2.8 million news articles and an additional 1 million short snippets (photo captions, video descriptions, etc.) published between 2001--2022 and collected from Voice of America's news websites.
We describe our process for collecting, filtering, and processing the data.
The source material is in the public domain, our collection is licensed using a creative commons license (CC BY 4.0), and all software used to create the corpus is released under the MIT License.
The corpus will be regularly updated as additional documents are published. 
\\
\\
}

\begin{document}

\maketitleabstract
\def\thefootnote{*}\footnotetext{Equal contribution.}\def\thefootnote{\arabic{footnote}}

\section{Introduction}
This work describes the first release of Multilingual Open Text (MOT), a collection of permissively licensed texts created with a goal of improving the amount of high-quality text available for lower-resourced languages.

MOT Release 1 consists of data collected from Voice of America (VOA) news websites.
Our broader goal is a corpus of open access multilingual text, and we plan to include data from other sources in future releases.
As part of the development of this corpus, we created infrastructure to continue to scrape new documents as they are published in order to provide subsequent releases with newly published and updated documents.
We have been using this infrastructure for several months to expand the corpus.
The corpus contains documents in many different languages, many of which are lower-resourced.

In this paper, we explain our process for collecting, filtering, and processing the data from VOA news websites in multiple languages and describe the resulting corpus. 
In Section~\ref{sec:relatedwork}, we motivate the need for this corpus and compare with similar lower-resourced language dataset creation efforts. 
In Section~\ref{sec:datasetdescription}, we describe the content of MOT. 
In Section~\ref{sec:datacollectionandprocessing}, we detail our process for creating the corpus. 
Finally, in Section~\ref{sec:limitationsfuturework}, we discuss limitations and future directions.
The corpus is available via GitHub.\footnote{\url{https://github.com/bltlab/mot/}}

\section{Related Work}
\label{sec:relatedwork}
A multilingual collection of unlabeled text can be useful for many tasks, especially for lower-resourced languages with limited freely-available text.
An unlabeled non-parallel corpus is typically the starting point for further annotation and dataset creation work.
Much of modern NLP relies on either pre-trained static or contextual word embeddings; in either case, these methods rely on large quantities of text data, which lower-resourced languages lack.

Even with the existence of multilingual Transformer models, like multilingual BERT \cite{devlin-etal-2019-bert} or XLM-R \cite{conneau-etal-2020-unsupervised}, unlabeled data from lower-resourced languages can be useful for adaptation of these models \cite{adelani-etal-2021-masakhaner,pfeiffer-etal-2020-mad}.
It is also possible to train a multilingual Transformer model without relying heavily on higher-resourced languages \cite{ogueji-etal-2021-small}.

There have been plenty of other works which have scraped news data for lower-resourced languages \cite{adelani-etal-2021-masakhaner,niyongabo-etal-2020-kinnews}.
\newcite{adelani-etal-2021-masakhaner} that also include partial scrapes of sections of VOA news sites.
\newcite{gezmu2021manually} used random samples of VOA news sites to create a spelling correction corpus for Amharic.
Unlike these data collection efforts, MOT intends to include a complete collection of VOA's documents rather than just enough data to meet the goals of a specific annotation effort.
Our resulting corpus also preserves metadata for each document which was discarded by other datasets.


There are a number of other existing resources that can be used as unlabeled data for lower-resourced languages.
The DARPA LORELEI program \cite{strassel-tracey-2016-lorelei,tracey-etal-2019-corpus,tracey-strassel-2020-basic} produced datasets for a number of lower-resourced languages. 
However, these datasets require payment or an LDC subscription which can be prohibitively expensive for speakers of those languages to access.
At the time of publication---over six years after the start of the program---many of the datasets planned for publication have not yet been released.

Many text collections for lower-resourced languages focus on parallel text for the purposes of machine translation.
The OPUS website hosts a number of parallel text datasets and related tools \cite{tiedemann-2012-parallel}. 
These parallel text datasets can also be treated as unlabeled monolingual text.

Among its many sources, OPUS contains data from the Christian Bible.
While the Christian Bible has been translated into more than 1,000 languages, it covers a very narrow domain that is not representative of most modern texts, is often translated into more archaic forms of each language, and reflects the perspective of its religious content.


JW300 \cite{agic-vulic-2019-jw300} is a corpus containing data in 300 different languages. 
It was extracted from jw.org, the website of the Jehovah's Witnesses (Watch Tower Bible and Tract Society of Pennsylvania).
While JW300 has been a useful resource for lower-resourced NLP, at the time of writing, it is not currently available due to it being distributed without permission from the copyright holders.
While we began work on MOT before JW300 became unavailable, the challenges of working with restrictively licensed source materials were one of the many factors that motivated us to create MOT.

There are also a number of multilingual corpora created from web-crawls such as Paracrawl \cite{espla-etal-2019-paracrawl,banon-etal-2020-paracrawl}, CC-aligned \cite{el-kishky-etal-2020-ccaligned},
WikiMatrix \cite{schwenk-etal-2021-wikimatrix}, and OSCAR \cite{ortiz-suarez-etal-2020-monolingual}. 

These web-crawled datasets tend to have a larger number of languages and larger numbers of documents. 
While OSCAR, for example, contains more documents and a higher number of languages, MOT contains data for some languages that OSCAR does not cover such as Cantonese, Dari, Hausa, Kinyarwanda, Lingala, Northern Ndebele, Oromo, Shona, and Tigrinya. 
Multilingual Open Text does not intend to compete with the size of these web-scraped corpora.
Instead, MOT aims to be a reliable scrape for particular established, edited, and permissively licensed data sources.
Web-scraped corpora can have issues with quality control as described in \newcite{caswell2021quality}.
While MOT covers fewer languages than many of these web-crawled corpora, it is more carefully curated and aims to avoid many of the pitfalls present in these larger-scaled corpora. 

MOT can also be used to build better language identification models to help create or improve larger scale corpora.

\section{Dataset Description}
\label{sec:datasetdescription}

\subsection{Source: Voice of America Online News}
\paragraph{Background.}
VOA was founded in 1942 and produces content for digital, television, and radio platforms in more than 40 languages.
It is the largest U.S. international broadcaster and has a weekly audience of an estimated 300 million people \cite{voa-code}. 
Because VOA's content is produced by employees of the United States government, it is in the public domain under U.S. federal law (\href{https://www.govinfo.gov/content/pkg/USCODE-2010-title17/html/USCODE-2010-title17-chap1-sec105.htm}{17 U.S.C. § 105}).
VOA's copyright statement in their terms of use also explicitly states that all content produced by VOA is in the public domain \cite{voa-terms}.

All documents not in the public domain were filtered out of this corpus.
The VOA copyright statement specifies that VOA has a license with the Associated Press (AP) to use AP content which is not in the public domain.
Although the VOA copyright statement does not explicitly mention them, we identified content written by Agence France-Presse (AFP) and Reuters appearing on VOA news websites.
We used automated methods to ensure that we did not include any articles from AP, AFP, and Reuters in our corpus.

\paragraph{Independent Journalism.}
Because VOA is funded by a government, it is worth discussing its independence as a news source and accordingly, the ethical considerations of using it in a corpus.
VOA maintains independence from U.S. political influences through the 1994 U.S. International Broadcasting Act, which prohibits any U.S. government official from interference in the objective reporting of news \cite{voa-firewall}. 
The VOA's journalistic code also requires accuracy, balance, fairness, and context in documents. For example, the code requires all staff who prepare content to not use negative terms to describe persons or organizations unless those individuals use those terms to describe themselves \cite{voa-code}.
These rules and standards suggest that the VOA operates independently, and thus a corpus derived from VOA content should be similar in its biases to corpora derived from other newswire sources, none of which are free of perspective or bias.

\subsection{Corpus Contents}
\label{sec:corpuscontents}

\begin{figure*}[tb]
\centering
\includegraphics[width=\linewidth]{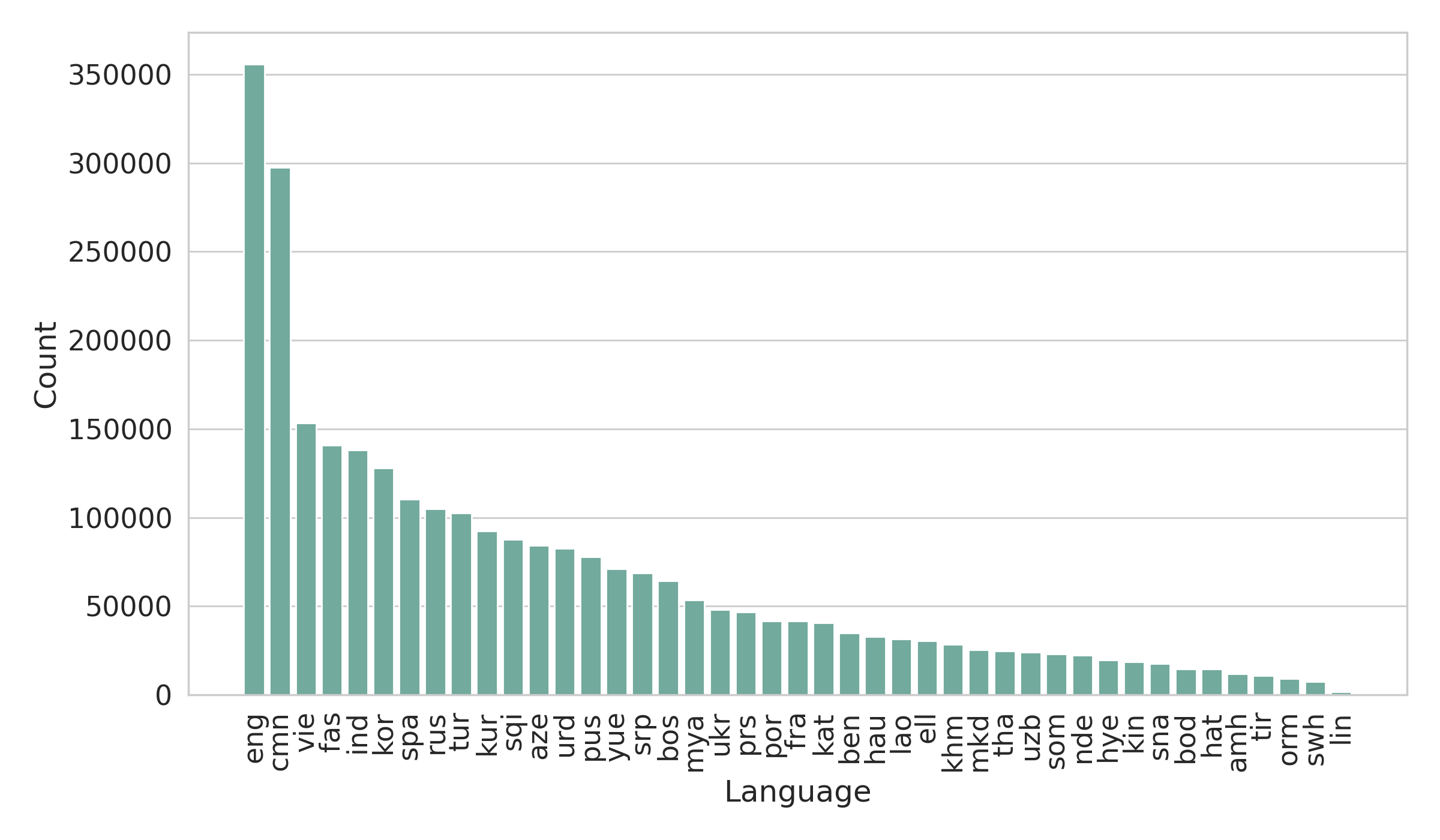}
\caption{Counts of news articles in MOT by language using ISO 639-3 codes}
\label{fig:counts-by-language}
\end{figure*}

This dataset contains paragraph-segmented data collected from 51 VOA news websites in the following 44 languages: Albanian, Amharic, Armenian, Azerbaijani, Bambara, Bangla, Bosnian, Burmese, Cantonese, Dari, English, French (African), Georgian, Greek, Haitian Creole, Hausa, Indonesian, Khmer, Kinyarwanda, Korean, Kurdish, Lao, Lingala, Macedonian, Mandarin Chinese, Northern Ndebele, Oromo, Pashto, Persian (Farsi), Portuguese (African), Russian, Serbian, Shona, Somali, Spanish, Swahili, Thai, Tibetan, Tigrinya, Turkish, Ukrainian, Urdu, Uzbek, and Vietnamese.
As noted, the French and Portuguese data is written primarily for African audiences.

\begin{figure}[tb]
\centering
\includegraphics[width=\linewidth]{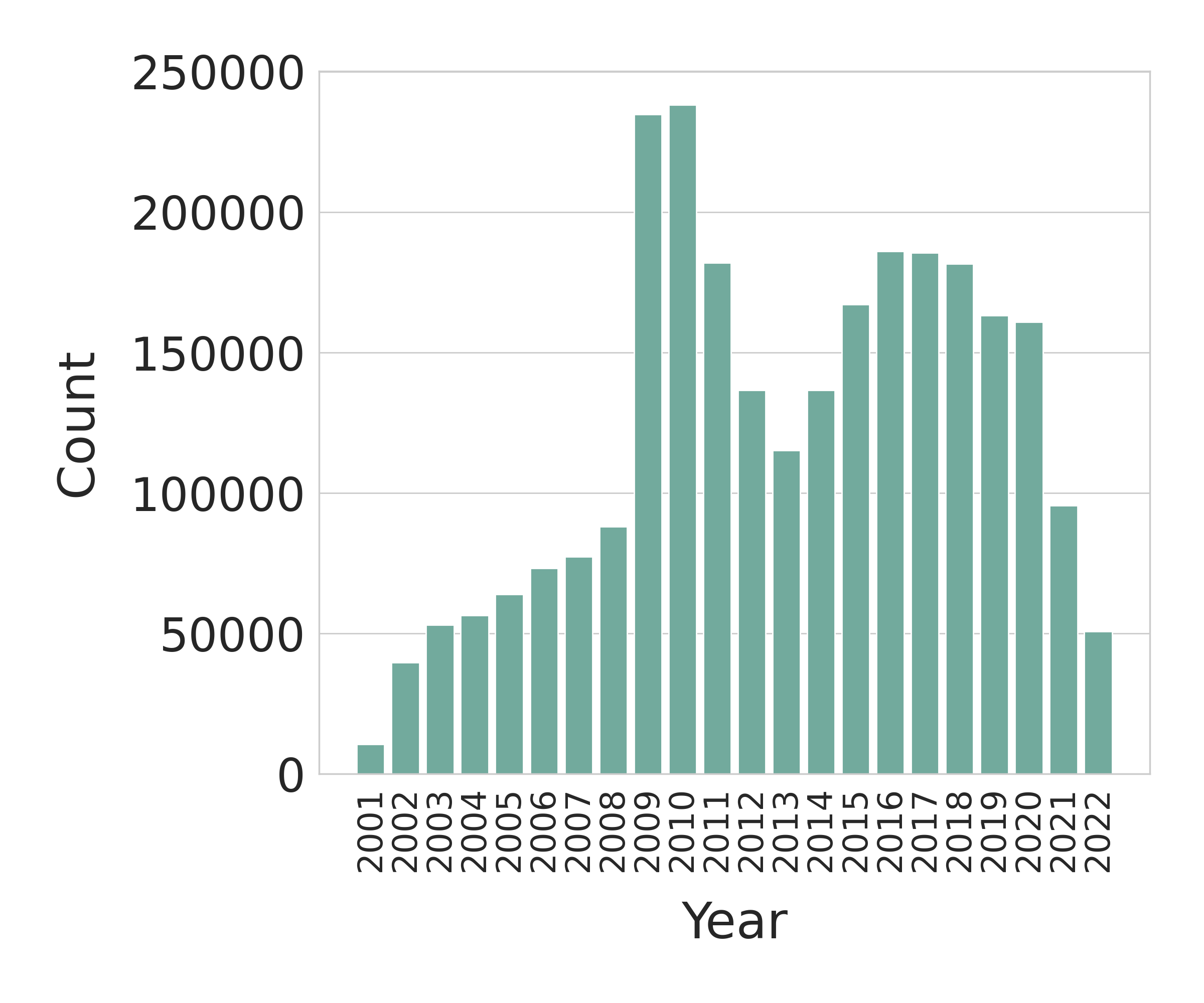} 

\caption{Counts of news articles in MOT by year}
\label{fig:counts-by-year}
\end{figure}

The counts of articles for each language are given in Figure~\ref{fig:counts-by-language}. 
While we have released the Bambara data for completeness, it contains essentially no articles, only short descriptions of other content (for example, photo captions, descriptions of audio stories, etc.).
This is largely due to how new the inclusion of Bambara is to VOA.\footnote{\url{https://www.insidevoa.com/a/6241315.html}} 
Currently the focus for the Bambara section of VOA is on radio and multimedia, not news articles.

As shown in Figure~\ref{fig:counts-by-year}, the corpus at the time of writing is comprised of articles published starting in 2001 up until May 1, 2022.\footnote{Documents with a timestamp prior to 2001 in the \texttt{time\_published} field were removed from the corpus. This includes 4 articles dated as 1899, 1900, 1997, and 1998 whose timestamps we believe to be incorrect.}
We do not know why there is such a large quantity of documents from 2010 or relatively few in 2021 compared to 2020; we suspect  these abnormalities have more to do with how content may be subdivided into documents than changes in the overall amount of content.
As more articles are published, we will continue to increase the size of the corpus.


\begin{table*}[tb]
\centering
\begin{tabular}{llrrrrr}
\toprule
Language & Code &   Article &   Audio &  Photo &   Video &       All \\
\midrule
Albanian            & sqi      & 87,854   & 4,986   & 230   & 16,326 & 109,396 \\
Amharic             & amh      & 11,990   & 9,429   & 220   & 1,818  & 23,457  \\
Armenian            & hye      & 19,671   & 0      & 63    & 6,938  & 26,672  \\
Azerbaijani         & aze      & 84,459   & 1,658   & 1,138  & 11,553 & 98,808  \\
Bambara             & bam      & 1       & 8,560   & 13    & 1,519  & 10,093  \\
Bangla              & ben      & 34,940   & 3,515   & 19    & 775   & 39,249  \\
Bosnian             & bos      & 64,267   & 7      & 457   & 10,192 & 74,923  \\
Burmese             & mya      & 53,456   & 9,540   & 467   & 18,309 & 81,772  \\
Cantonese           & yue      & 71,162   & 19,433  & 438   & 16,378 & 107,411 \\
Dari                & prs      & 46,885   & 18,423  & 239   & 6,334  & 71,881  \\
English             & eng      & 355,821  & 188,143 & 1,561  & 8,594  & 554,119 \\
French (African)    & fra      & 41,570   & 38,491  & 491   & 10,796 & 91,348  \\
Georgian            & kat      & 40,572   & 5,905   & 294   & 5,762  & 52,533  \\
Greek               & ell      & 30,444   & 134    & 26    & 64    & 30,668  \\
Haitian Creole      & hat      & 14,478   & 8,812   & 300   & 7,012  & 30,602  \\
Hausa               & hau      & 32,957   & 15,873  & 1,167  & 2,647  & 52,644  \\
Indonesian          & ind      & 138,121  & 83,092  & 1,400  & 17,786 & 240,399 \\
Khmer               & khm      & 28,468   & 8,916   & 476   & 3,161  & 41,021  \\
Kinyarwanda         & kin      & 18,697   & 10,322  & 271   & 503   & 29,793  \\
Korean              & kor      & 127,867  & 1,546   & 304   & 1,108  & 13,0825 \\
Kurdish             & kur      & 92,335   & 15,640  & 1,614  & 7,316  & 116,905 \\
Lao                 & lao      & 31,619   & 3,519   & 230   & 943   & 36,311  \\
Lingala             & lin      & 1,744    & 3,026   & 16    & 1,471  & 6,257   \\
Macedonian          & mkd      & 25,500   & 2      & 94    & 4,775  & 30,371  \\
Mandarin            & cmn      & 297,587  & 37,060  & 1,269  & 16,977 & 352,893 \\
Northern Ndebele    & nde      & 22,516   & 5,530   & 43    & 3,379  & 31,468  \\
Oromo               & orm      & 9,225    & 324    & 82    & 513   & 10,144  \\
Pashto              & pus      & 77,769   & 46,526  & 313   & 16,685 & 141,293 \\
Persian             & fas      & 140,724  & 0      & 1     & 0     & 140,725 \\
Portuguese (African) & por      & 41,620   & 4,266   & 458   & 6,170  & 52,514  \\
Russian             & rus       & 104,817  & 631    & 456   & 12,507 & 118,411 \\
Serbian             & srp      & 68,805   & 173    & 164   & 6,476  & 75,618  \\
Shona               & sna      & 17,594   & 7,589   & 10    & 2,858  & 28,051  \\
Somali              & som      & 23,192   & 14,788  & 194   & 202   & 38,376  \\
Spanish             & spa      & 110,428  & 3,880   & 44    & 2,090  & 11,6442 \\
Swahili             & swh      & 7,388    & 10,361  & 458   & 5,697  & 23,904  \\
Thai                & tha      & 24,732   & 7,930   & 133   & 1,278  & 34,073  \\
Tibetan             & bod      & 14,715   & 22,964  & 4     & 7,719  & 45,402  \\
Tigrinya            & tir      & 10,774   & 2,359   & 182   & 1,094  & 14,409  \\
Turkish             & tur      & 102,560  & 168    & 745   & 17,560 & 121,033 \\
Ukranian            & ukr      & 48,297   & 25     & 639   & 16,963 & 65,924  \\
Urdu                & urd      & 82,642   & 3,540   & 2,459  & 12,724 & 101,365 \\
Uzbek               & uzb      & 24,129   & 6,676   & 2,736  & 10,083 & 43,624  \\
Vietnamese          & vie      & 153,287  & 7,594   & 674   & 20,811 & 182,366 \\
\midrule
Total &   & 2,837,679 & 641,356 & 22,592 & 323,866 & 3,825,493 \\
\bottomrule
\end{tabular}
\caption{Counts of documents by content type and ISO 639-3 codes for each language included in MOT.}
\label{table:contenttype-counts}
\end{table*}

The corpus is organized by VOA site and further organized by content type.
Some languages in VOA are further divided into separate domains.
For example, English includes VOA News (global news), VOA Zimbabwe, Editorials, and an English Learning site. 
Pashto, Kurdish, and Khmer also have more than one domain, where the distinction is typically a differing region or dialect (for example, Sorani and Kurmanji for Kurdish).
The content types that we encountered in VOA pages' metadata were as follows: article, audio, video, photo, poll, quiz, index, author, schedule, subscribe, upload, account, and comment. 

We focus on extracting data of content type article,\footnote{Example: \url{https://www.voanews.com/a/2020-usa-votes_bidens-cabinet-picks-include-some-firsts/6198990.html}} which is a typical news article. 
However, we also include audio, video, and photo pages as they contain some usable text data in the form of titles, short captions, or descriptions.
The content types audio\footnote{Example: \url{https://www.voanews.com/t/60.html}} and video\footnote{Example: \url{https://www.voanews.com/a/episode_nuclear-power-cautiously-embraced-bidens-green-goals-4711476/6117084.html}} includes documents associated with audio and video media.
The content type photo\footnote{Example: \url{https://www.voanews.com/a/2808902.html}} includes documents that mainly include a series of captioned images.
The counts of documents in each content type can be seen in Table~\ref{table:contenttype-counts}\footnote{These reflect our best counts of the data, but these change regularly as new data is scraped and data issues are addressed.}. 
Most languages have more content of type article, yet some languages, like Swahili, may have more of a focus on radio, and thus contain more audio files. 

For some languages, we have few or no documents for certain content types like audio or video. 
This is typically not because there is no audio or video in that language, but because the audio and video in that language did not contain captions from which to extract text data, the captions were so short that they were unlikely to represent meaningful content, or the captions were in an unexpected format that caused our extraction to miss it. 
Pages of content type poll, quiz, index, author, schedule, subscribe, upload, account, and comment are not included in our final data release.
These typically contained little or no data, were more complicated to extract from, or in the case of index, duplicated descriptions from other pages where we were able to perform more complete extractions.

All content provided in this corpus is text, so for media like photos and videos, the data is the text description or a caption; it is not extracted from the media itself.
Paragraph breaks from the original HTML are preserved and  
documents are represented as lists of paragraphs, which contain lists of sentences, which contain lists of tokens.

File names sometimes contain abbreviated headlines, but occasionally the headline used for the file name is in a different language than the actual headline and text appearing in the document. 
This is likely the result of editorial errors and may reflect that the document was adapted or translated from a document in another language.
Each VOA domain is provided as a separate .tgz file in our release with subdirectories for different content types like article, audio, video, etc.

All languages are identified using ISO 639-3 codes.
Each file contains the following fields:
\setlist{nolistsep}
\begin{itemize}[noitemsep]
    \item \texttt{filename}: the name of the file derived from the URL
    \item \texttt{url}: the URL from which the document was retrieved
    \item \texttt{url\_origin}: the sitemap from which the URL was retrieved
    \item \texttt{content\_type}: the type of content (e.g., article, audio, photo, video) of the document
    \item \texttt{site\_language}: the language of the VOA site
    \item \texttt{time\_published}: the timestamp for when the document was published
    \item \texttt{time\_modified}: the timestamp for when the document was last modified
    \item \texttt{time\_retrieved}: the timestamp for when the document was retrieved from the sitemap
    \item \texttt{title}: the title of the document
    \item \texttt{authors}: the author(s) of the document
    \item \texttt{paragraphs}: the text extracted from the document
    \item \texttt{n\_paragraphs}: the number of paragraphs in the document
    \item \texttt{n\_chars}: the number of characters in the document
    \item \texttt{cld3\_detected\_languages}: the language(s) identified by CLD3 from the full extracted text of the document (see Section~\ref{sec:langid})
    \begin{itemize}
        \item \texttt{language}: the language outputted by CLD3
        \item \texttt{probability}: the probability that the language identified is correct (passed directly from CLD3)
        \item \texttt{is\_reliable}: if probability is above 0.7 (passed directly from CLD3)
        \item \texttt{proportion}: the proportion of the text identified as the language (passed directly from CLD3)
    \end{itemize}
    \item \texttt{predicted\_language}: the language that we predict that the document is in, based on rules that take into account the site, the CLD3 predictions, and whether the site language is supported by CLD3
    \item \texttt{keywords}: the terms relating to the text content of the document
    \item \texttt{section}: the subpage the document falls under
\end{itemize}

These additional fields are included only for subset of languages:
\setlist{nolistsep}
\begin{itemize}[noitemsep]
    \item \texttt{sentences}: the text extracted from the document segmented into sentences
    \item \texttt{n\_sentences}: the number of sentences in the document
    \item \texttt{tokens}: the text extracted from the document segmented into tokens
    \item \texttt{n\_tokens}: the number of tokens in the document
    \item \texttt{parallel\_english\_article}: the URL for the English document from which the current document was translated from into the site language (this currently only appears in Lao articles)
\end{itemize}

\subsection{How Low Resourced?}
\label{sec:taxonomy} 
While there is no single way of classifying lower-resourced languages due to the large number of intersecting factors that contribute to such a designation, 
\newcite{joshi-etal-2020-state} created a taxonomy of resource availability for languages based on the amount of labeled and unlabeled data.
The scale goes from 0 (lowest resources) to 5 (highest resources).
Although the taxonomy is an oversimplification of the state of resources for a language since there are many more dimensions (domain, task, medium, register, etc.) by which data can be categorized, it can still provide some sense of low-resourced-ness.

Of the 44 languages included MOT Release 1, only 4 are considered ``winners'' at level 5.
16 of the languages are classified as level 1, ``scraping-bys,'' which is described as having essentially no labeled data and very little existing unlabeled data.
MOT also includes 3 languages classified as level 0, ``left-behinds,'' Haitian Creole, Northern Ndebele, and Dari.

Another way of evaluating the low-resourced-ness of MOT is to compare with Wikipedia. 
Because Wikipedia is a commonly used resource for multilingual text, languages that have poor representation in Wikipedia could be considered more lower-resourced.
We compare MOT articles to Wikipedia articles by counts of characters in each dataset in Table~\ref{tab:mot-wikipedia}.
We use character counts since tokens are dependent on the quality of the tokenizer and lower-resourced languages may not have adequate tokenization.
As seen in Table~\ref{tab:mot-wikipedia}, MOT contains more data than Wikipedia in 13 languages, demonstrating MOT's potential value in providing more unlabeled text data for lower-resourced languages.

While it is true that the highest-resourced languages such as English or French contained in MOT initially do not appear to be much of a contribution when plenty of resources exist for these languages, we include them for completeness and because much of the text in the VOA documents has a regional focus that may not be present in existing  datasets. 

For example, portions of the English data focus on news in Zimbabwe while a portion of the Portuguese data is centered around Mozambique. 
This can matter for annotation projects that may wish to use monolingual data that is region-specific.\footnote{As an example, one early adopter of our corpus wished to translate news data focused on Mozambique from Portuguese into eMakhuwa to create a parallel corpus.}
While there is existing Mozambique-focused Portuguese data available from \newcite{davies2006corpus}, we are not aware of any usable data for Zimbabwe-focused English. 
We were able to identify one corpus focusing on Zimbabwe English textbooks, but as it was stored on magnetic tapes,  we were not able to locate a copy \cite{Louw1993CORPUSOZ}.

\begin{table}[tb]
\centering
\begin{tabular}{l*{6}r}
\toprule
Lang. & Wikipedia & MOT \\
\midrule
hau & 37,141,190  & 38,341,381  \\
khm & 34,048,132  & 93,948,921  \\
kin & 3,822,464   & 23,881,242  \\
lao & 5,999,270   & 61,419,714  \\
lin & 1,502,089   & 1,744,378   \\
nde & 0           & 31,600,251  \\
orm & 2,257,827   & 10,469,043  \\
prs & 0           & 67,421,867 \\
pus & 37,683,579  & 127,570,695 \\
sna & 6,606,352   & 29,132,817  \\
som & 12,018,769  & 18,244,956  \\
sqi & 157,576,602 & 181,583,961 \\
tir & 152,456     & 7,645,809   \\
\bottomrule
\end{tabular}
\caption{Counts of characters in Wikipedia and MOT for lower-resourced languages where MOT provides a higher count}
\label{tab:mot-wikipedia}
\end{table}

\section{Data Collection and Processing}
\label{sec:datacollectionandprocessing}
\subsection{Scraping VOA}

While this work is not the first to scrape text data from Voice of America, it is to the best of our knowledge the most thorough and complete scraping effort of the text contained on the Voice of America collection of websites.
The data collection process starts with manually creating a list of all the different VOA domains along with their ISO 639-3 language codes.
We then use the list of VOA domains to automatically get all the URLs from each site's sitemap.
The current release includes documents retrieved from sitemaps between June 16, 2021 and May 1, 2022.

When scraping a page, we extract the title, description, keywords, and author(s) from the HTML meta tags. 
We also attempt to collect the canonical link, publication date, modification date, and content type for each page. 
In addition to the sitemaps, we used the Internet Archive's Wayback CDX Server API\footnote{\url{https://github.com/internetarchive/wayback/blob/master/wayback-cdx-server/README.md}} to collect URLs for each domain.
Of the URLs we retrieved using the Internet Archive, the vast majority were duplicates. 
In the case where the pages were of content type article, only 5 Thai pages and only 3 French pages were not already retrieved through the sitemaps.
While this process of using the Internet Archive in addition to the sitemaps did not produce meaningful gains in content, it did help us to verify that we are not missing any easily retrievable content from the sitemaps.

The scraped pages are maintained in a database and we compare against existing pages' URLs and canonical links in order to de-duplicate and use the most recent version of a page. 
Our collection effort of VOA data differs from other efforts in that we regularly do an updated scrape. 
We have scraped periodically\footnote{Re-scraping occurs roughly once a month.} since beginning our collection effort in summer 2021.
The gains in numbers of previously unseen URLs in roughly a month's time varies from a few hundred to about 2,000 for languages other than English. 
The Greek section of VOA is no longer being updated, so there are never new URLs for that section. 
We also notice some URLs are no longer found in the sitemaps between our scraping efforts; however, the number of URLs lost is quite small.

For example, only 720 URLs went missing in Persian between December 1, 2021 and January 1, 2022, which is relatively small compared to the 141,060 documents we extracted. 
For the same time period, 25 languages had not lost any URLs in the sitemaps. 
We can also report anecdotally that many of these lost URLs are either video clips with little or no caption content or are sites that were updated and have a newer URL, which we attempt to de-duplicate if a canonical link was present.

\subsection{Extracting Text from HTML Documents}
We now turn to the process of extracting text data from the raw HTML scraped from VOA.
All relevant text content from each document is extracted and paragraph breaks from the HTML are maintained in the output. 
However, not all data that is extracted from paragraph tags or the usual div tags is actually part of the document content.  We remove repetitive and meaningless content, such as user comments and sentences that consist of the equivalent of ``login.''
If the page contains no valid text, it is not included in the output.

The \texttt{filename} we create is derived from the URL and includes everything following the top-level domain. 
If the name of the file is too long, the \texttt{filename} is shortened to only the last 100 characters.


\subsection{Language Identification and Filtering}
\label{sec:langid}
Not all of the documents in VOA are consistently in one language. 
While code switching exists, most of the mixed language use that we observed in the corpus were sentences that were translations of other content in the document rather than instances of natural code switching.
Unfortunately, these translated portions of such documents did not appear to be systematic enough to extract parallel text in most cases.
In some cases, this is because the document is a translation, but the captions remain in the reported language. 
In other cases, the document may contain the English translation or may be a part of VOA's language learning site that was miscategorized. 
We attempt to filter out heavily multilingual text along with documents that erroneously contain mostly English despite claiming to be written in another language.

\paragraph{CLD3.}
We use CLD3 for our language ID in the filtering process. 
Compact Language Detector version 3 (CLD3) \cite{cld3} is a neural network model for language identification that supports over 100 languages and scripts. 
The model outputs the languages identified as BCP-47-style language codes, along with its corresponding probability, reliability, and proportion (see Section~\ref{sec:corpuscontents} for more information about these fields).
CLD3 does not support the following languages in MOT: Azerbaijani, Bambara, Cantonese, Dari, Kinyarwanda, Lingala, Northern Ndebele, Oromo, Tibetan, and Tigrinya.
Because these languages are unsupported, we do not use the language ID predictions for our \texttt{predicted\_language} field and instead rely on VOA's reported language based on which domain the site is from.
We do include the main CLD3 prediction information, but end users should take note that certain languages are likely to be misrecognized. 
For example, Tigrinya is regularly classified as Amharic by CLD3 since it is not supported. 

\paragraph{Filtering Process.}
CLD3 was used to identify the language present in the extracted text with a maximum of 5 languages.
This was used to determine the predicted language of the document.
We filter at the paragraph level and at the document level. 
At the paragraph level, we filter only for confidently English paragraphs in non-English sections of VOA. 
If the probability is greater than 0.7 and the proportion of the paragraph is more than 0.25, the English paragraph is discarded.
Because URLs in text tend to get identified as English by CLD3, this also helps to filter out URLs. 
This paragraph level filtering is useful as there are some documents that will be almost entirely in one language with just one or a few paragraphs in English. Typically, these paragraphs in English are also redundant with the main language of the document.\footnote{\url{https://www.voaswahili.com/a/netanyahu-aipongeza-marekani-kwa-usimamizi-wa-kurejesha-mahusiano-kati-ya-israeli-na-sudan/5634218.html}}
It is also common for the English contamination to be a translation of just a few quotes in the document.\footnote{\url{https://www.voaswahili.com/a/ndege-ya-ethiopian-airlines-imeanguka-na-juhudi-za-kuitafuta-zaendelea/4822036.html}}

At the document level, we also run language ID on the original text before paragraph level filtering. 
If CLD3 is confident in one language, the predicted language is assumed to be either the original sitemap language or English as CLD3 does not predict all of the languages encountered in the corpus. 
If CLD3 is confident that the majority of the document is either English in a non-English section, or non-English in an English section, the document is filtered out.
If CLD3 has identified multiple languages with a probability above 0.9 and a proportion above 0.05, the predicted language is listed as ``mul.''
All documents include a prediction of the language expected from the output of CLD3. 
Every document is predicted to be written in the site language unless CLD3 has identified more than one language from the text (``mul'') or CLD3 has identified only English present in the document (``eng''), in which case the document is not included.



\subsection{Sentence Segmentation and Tokenization}
\label{sec:segmentation}

\begin{table}[tbh]
\centering
\resizebox{\linewidth}{!}{
\begin{tabular}{lrrr}
\toprule
Language &  Documents &  Sentences &      Tokens \\
\midrule
amh      & 23,457    & 91,051    & 1,960,739   \\
aze      & 98,808    & 644,156   & N/A        \\
bos      & 74,923    & 577,114   & N/A         \\
cmn      & 352,893   & 2,177,530  & 130,214,418 \\
ell      & 30,668    & 155,284   & 6,090,066   \\
eng      & 554,119   & 4,537,686  & 193,129,912 \\
fas      & 140,725   & 871,887   & 35,929,609  \\
fra      & 91,348    & 507,058   & 19,966,932  \\
hat      & 30,602    & 100,558   & N/A         \\
hau      & 52,644    & 244,043   & N/A         \\
hye      & 26,672    & 150,086   & 5,064,730   \\
ind      & 240,399   & 1,245,778  & 38,383,509  \\
khm      & 41,021    & 392,724   & 19,027,222  \\
kin      & 29,793    & 119,298   & N/A       \\
kor      & 130,825   & 1,516,790  & 41,548,365  \\
lao      & 36,311    & 532,944   & 12,058,686  \\
lin      & 6,257     & 18,757    & N/A       \\
mkd      & 30,371    & 245,127   & N/A         \\
mya      & 81,772    & 657,459   & 36,006,802  \\
nde      & 31,468    & 211,156   & N/A         \\
orm      & 10,144    & 57,187    & N/A         \\
por      & 52,514    & 427,612   & 13,864,438  \\
prs      & 71,881    & 461,203   & 14,633,719  \\
pus      & 141,293   & 838,726   & N/A         \\
rus      & 118,411   & 1,051,201  & 51,451,892  \\
sna      & 28,051    & 189,093   & N/A         \\
som      & 38,376    & 131,501   & N/A         \\
spa      & 116,442   & 911,685   & 33,352,028  \\
sqi      & 109,396   & 793,622   & N/A         \\
srp      & 75,618    & 618,884   & 26,544,508  \\
swh      & 23,904    & 63,761    & N/A         \\
tha      & 34,073    & 262,953   & 9,428,506   \\
tir      & 14,409    & 76,283    & 1,784,820   \\
tur      & 121,033   & 861,882   & 31,419,370  \\
ukr      & 65,924    & 363,540   & 17,232,122  \\
urd      & 101,365   & 986,220   & 40,805,126  \\
uzb      & 43,624    & 314,141   & N/A         \\
vie      & 182,366   & 1,138,882  & 59,843,930  \\
yue      & 107,411   & 70,1411   & 34,730,065  \\
\midrule
   Total &  3,561,311 & 25,246,273 & 874,471,514 \\
\bottomrule
\end{tabular}
}
\caption{Counts of documents, sentences, and tokens for languages with sentence segmentation and tokenization}
\label{table:sentencetoken-counts}
\end{table}

\paragraph{Segmentation.}
We primarily use Ersatz \cite{wicks-post-2021-unified} for sentence segmentation; however, off-the-shelf monolingual models provided for Ersatz do not cover all of the languages in MOT. 
We attempted to use the multilingual model provided by Ersatz, but it had unsatisfactory performance in some languages. 
In Swahili, it failed to segment the abbreviation for \emph{doctor}, \emph{Dkt.} correctly. 
We also noticed some instances of periods after first initials being treated as sentence boundaries in Greek, likely because Ersatz was not trained on any language using the Greek alphabet.
It also did not contain any Ge'ez script punctuation as candidates for sentence splits and was therefore unusable for Amharic or Tigrinya. 
Thai and Lao, which do not have sentence ending punctuation, also created challenges.
Because the multilingual segmentation model had sub-optimal performance for languages it was not trained on, we have chosen only to release sentence breaks and tokenization for those languages where we could provide more reliable segmentation.

We used PyThaiNLP \cite{pythainlp} for Thai and \texttt{amseg} \cite{fi13110275} for Amharic and Tigrinya. 
\texttt{amseg} is a rule-based Amharic segmenter, but as it is based on whitespace and Ge'ez script punctuation, we used it for Tigrinya in addition to Amharic. 
Parsivar \cite{mohtaj-etal-2018-parsivar} was used for Persian, \texttt{khmer-nltk} for Khmer, LaoNLP\footnote{\url{https://github.com/wannaphong/LaoNLP}} for Lao, and \texttt{razdel} for Russian. 
We also use Stanza \cite{qi2020stanza} for Armenian, Burmese, Greek, Indonesian, Korean, Portuguese, Serbian, Ukrainian, Urdu, and Vietnamese. 
We trained custom Ersatz models using paragraph breaks from MOT for the remaining languages.

As \newcite{wicks-post-2021-unified} point out, there tends to be a lack of reliable test sets for sentence segmentation, so we have not yet independently vetted the performance of these segmenters.
For languages in which we do not yet have satisfactory sentence segmentation, we do not provide sentence breaks. 
In Table~\ref{table:sentencetoken-counts}, we provide counts of sentences and tokens for the languages where we are able to provide segmentation and tokenization.

\paragraph{Tokenization.}
We used spaCy \cite{spacy2} for tokenization in English, Cantonese, French, Mandarin Chinese, Russian, Spanish, and Turkish. 
PyThaiNLP \cite{pythainlp} is used to tokenize Thai, and \texttt{amseg} \cite{fi13110275} to tokenize Amharic and Tigrinya. \texttt{khmer-nltk} \cite{hoang-khmer-nltk} was used for Khmer tokenization. 
Stanza \cite{qi2020stanza} is also used for tokenization in the same languages it is used for sentence segmentation.
 We hope to provide more robust tokenization and segmentation in future releases.

\section{Limitations and Conclusion}
\label{sec:limitationsfuturework}
Extracting text from HTML from a complex network of sites like VOA is non-trivial, and although we have done our best to ensure complete, clean extractions, we expect users of this resource will discover issues.

There are still a number of languages where we do not have reliable sentence segmentation and tokenization.
We would like to improve language identification to better identify documents with multiple languages, as CLD3 does not cover all of the languages in MOT.
We plan to continue to increase the size of the corpus as VOA publishes more documents, and we plan to expand MOT by adding other permissively-licensed texts to expand our coverage of lower-resourced languages.

There are many ways in which MOT could be used in future work. 
For lower-resourced languages, MOT provides a valuable source of high-quality unlabeled text, and it could be used with minimal annotation effort to train language identification, sentence segmentation, and tokenization systems.
Sections of MOT could also be used for annotation projects to create labeled data for tasks like document classification, named entity recognition, and syntactic or semantic parsing.

Because MOT includes publication time metadata, it may be possible to use MOT to create semi-parallel text. 
While we do not include audio or images as part of our release, others may want to make use of the included source URL and employ the captions on the photo content type for image captioning in lower-resourced languages. 

We have presented a new corpus containing unlabeled text data in 44 languages, many of them lower-resourced languages for which this represents a substantial increase in the amount of available text data.
The data in this corpus is in the public domain, and the corpus is positioned to grow in future releases as new documents are published. 
We look forward to the opportunity to further refine the extraction and increase the usefulness of MOT as speakers of the languages contained in it begin to make use of it.

\section{Acknowledgments}
We thank the early adopters of our resource---many of whom are members of the Masakhane community---who used preliminary releases and offered feedback.
This work was supported by a 2021 Brandeis University Provost Research Grant.

\section{Bibliographical References}

\bibliographystyle{lrec2022-bib}
\bibliography{main}

\end{document}